\def\year{2022}\relax
\documentclass[letterpaper]{article}
\usepackage{times} 
\usepackage{helvet} 
\usepackage{courier} 
\usepackage[hyphens]{url} 
\usepackage{graphicx} 
\usepackage{graphicx}  
\usepackage{natbib}  
\usepackage{caption}  
\DeclareCaptionStyle{ruled}%
 {labelfont=normalfont,labelsep=colon,strut=off}
\usepackage[utf8]{inputenc} 
\usepackage[T1]{fontenc}    
\usepackage{hyperref}       
\usepackage{url}            
\usepackage{booktabs}       
\usepackage{amsfonts}       
\usepackage{nicefrac}       
\usepackage{microtype}      
\usepackage{graphicx}
\usepackage{todonotes}
\usepackage{wrapfig}
\usepackage{enumitem}

\include{math}
\usepackage{natbib}

\usepackage{xcolor}

\usepackage{amsthm}
\usepackage{amssymb,amsmath,amsthm,amsfonts}

\newtheorem{theorem}{Theorem}
\newtheorem{proposition}{Proposition}
\newtheorem{lemma}{Lemma}

\theoremstyle{definition}
\newtheorem{definition}{Definition}[section]

\usepackage{tikz}
\usetikzlibrary{positioning,arrows}
\usetikzlibrary{decorations.markings}
\usetikzlibrary{calc}

\tikzstyle{vecArrow} = [thick, decoration={markings,mark=at position
   1 with {\arrow[semithick]{open triangle 60}}},
   double distance=1.4pt, shorten >= 5.5pt,
   preaction = {decorate},
   postaction = {draw,line width=1.4pt, white,shorten >= 4.5pt}]
\tikzstyle{innerWhite} = [semithick, white,line width=1.4pt, shorten >= 4.5pt]

\tikzstyle{every picture}+=[remember picture]
\tikzstyle{state}=[shape=circle,draw=blue!50,fill=blue!20, minimum size=1.3cm]
\tikzstyle{unity}=[shape=circle,draw=blue!50, minimum size=1.3cm]
\tikzstyle{observation}=[shape=rectangle,draw=orange!50,fill=orange!20]
\tikzstyle{lightedge}=[<-,dotted]
\tikzstyle{mainstate}=[state,thick]
\tikzstyle{mainedge}=[<-,thick]
\tikzstyle{revmainedge}=[->,thick]
\tikzset{onslide/.code args={<#1>#2}{%
  \only<#1>{\pgfkeysalso{#2}} 
}}

\tikzset{
  treenode/.style = {align=center, inner sep=0pt, text centered,
    font=\sffamily},
  arn_n/.style = {treenode, circle, font=\sffamily\bfseries, draw=black,
    text width=2.1em},
  arn_r/.style = {treenode, circle, black, draw=black, fill=red,
    text width=2.1em, very thick},
  arn_x/.style = {treenode, rectangle, draw=black,
    minimum width=1.5em, minimum height=2.5em},
  arn_s/.style = {treenode, rectangle, draw=black,
    minimum width=0.5em, minimum height=0.5em}
}

\tikzstyle{highlight}=[red,ultra thick]

\title{Trees in Transformers: \\ A Theoretical Analysis of the Transformer's Ability to Represent Trees}

\author{
    Qi He, \textsuperscript{\rm 1} Jo\~ao Sedoc, \textsuperscript{\rm 2} Jordan Rodu \textsuperscript{\rm 3}
}

\date{}

\begin{document}

\maketitle{}

\begin{abstract}

Transformer networks \citep{Vaswani2017AttentionIA} are the de facto standard architecture in natural language processing. To date, there are no theoretical analyses of the Transformer's ability to capture tree structures. We focus on the ability of Transformer networks to learn tree structures that are important for tree transduction problems. We first analyze the theoretical capability of the standard Transformer architecture to learn tree structures given enumeration of all possible tree backbones, which we define as trees without labels. We then prove that two linear layers with ReLU activation function can recover any tree backbone from any two nonzero, linearly independent starting backbones. This implies that a Transformer can learn tree structures well in theory. We conduct experiments with synthetic data and find that the standard Transformer achieves similar accuracy compared to a Transformer where tree position information is explicitly encoded, albeit with slower convergence. This confirms empirically that Transformers can learn tree structures.

\end{abstract}

\let\thefootnote\relax\footnotetext{
    \textsuperscript{\rm 1} New York University, qh2133@nyu.edu, \textsuperscript{\rm 2} New York University, jsedoc@stern.nyu.edu,
    \textsuperscript{\rm 3} University of Virginia, jsr6q@virginia.edu}

\section{Introduction}
\label{sec:intro}

Neural models recently garnered widespread attention for their state-of-the-art (SoTA) performance gains. For instance, Transformer networks \citep{Vaswani2017AttentionIA}, an innovation with a simple and parallelizable architecture, achieved SoTA results in neural machine translation \citep{edunov2018understanding} and subsequently many other NLP tasks~\citep{brown2020language}. Empirical results suggest that Transformers are able to capture syntactic dependencies in natural language analysis~\citep{hewitt2019structural}, but to the best of our knowledge, there is no theoretical analysis specifically about this property of Transformers.

In this paper, we address the following questions: can Transformers capture tree structures (\autoref{sec:initialapproach}); What are the limits of that ability (i.e. how large does the Transformer need to be and how deep a tree can a Transformer network handle (\autoref{sec:initialapproach} and \autoref{sec:secondapproach})); How can we make the learning process more efficient (\autoref{sec:experiments}). Given the widespread use of Transformer networks in speech and natural language processing, these questions can help to answer what size of networks are needed as well as start to give guidance as to when explicitly providing tree position encodings is important. Our experiments use synthetic data as we aim to clearly identify and answer these questions agnostic of the application area.

Our paper contributes to the understanding of the ability of Transformers to learn tree structures in the following ways. First, we theoretically examine the Transformer's ability to learn tree structures with only sequential positional encoding. 
Then, we investigate whether the Transformer can recover tree structures given only a small number of observed trees in the training data. 
Next, we validate the theoretical result with empirical experiments using synthetic data where we encode sequential information and compare this with the upper bound with a modified Transformer that explicitly encodes tree structure information. This allows us to evaluate how well standard Transformer networks can learn tree structures. 
Finally, we conclude with an analysis of the theoretical versus empirical results. 

\section{Transformers Can Learn Tree Structures}
\label{sec:initialapproach}

We start by asking if a 1-layer Transformer is able to find the structure of a tree (i.e. all information in the tree except node values), up to a maximum pre-specified sequence length. 
To show this is possible, we explicitly construct such a Transformer in this section.

Following the notations from \citet{Vaswani2017AttentionIA}, we first describe a 1-layer 1 attention head Transformer encoder. Let $\mathbf{z}$, the serialized input tree, be an $n$-dimensional vector, with padding applied when necessary. We can embed each element of $\mathbf{z}$ in $d$-dimensional space and collect these embeddings in the rows of a matrix $X_0\in \mathbb{R}^{n\times d}$. Let $P\in \mathbb{R}^{n\times d}$ be the positional encoding matrix. Then the output of the embedding layer is
\[X_1 = X_0+P\in \mathbb{R}^{n\times d}.\]

The output of the attention layer is
\[X_2=LayerNorm(X_1 + \text{Attention}(Q,K,V))\in\mathbb{R}^{n\times d}\] where 
\[\text{Attention}(Q,K,V) = \text{softmax}\Big(\frac{QK^T}{\sqrt{d}}\Big)V,\] \[Q=X_1W^Q, ~~K=X_1W^K,~~ V=X_1W^V\]
and LayerNorm is layer normalization operation from \citep{ba2016layer}. The trainable parameters of the attention layer are $W^Q,W^K,W^V\in \mathbb{R}^{d\times d}$.

Finally, let $ReLU(x)=\max(0,x)$. Each row of $X_2$, denoted as $x\in\mathbb{R}^n$, goes through a feed-forward layer, 
$$X_3=LayerNorm(FFN(x)+x),$$ where
\[FFN(x)=ReLU(xW_1+b_1)W_2+b_2, \tag{2.1} \label{eq:relu} \]
with trainable parameters $W_1,W_2\in\mathbb{R}^{d\times d}$ and $b_1,b_2\in\mathbb{R}^{d}$.

\subsection{Finding the backbone of a tree}
In order to describe tree structures, we define tree backbone and backbone vectors. 

\definition{The backbone of a tree $T$ is the rooted unlabelled tree with the same tree structure as $T$.}

In this way, we are actually calling the structure of a tree its tree backbone. As it is necessary to vectorize backbones, we need a backbone function that maps each backbone to a vector.

\definition{Given a serialization of trees with maximum length $n$, a backbone function is an element-wise function $\mathbf{f}:\Sigma^n\rightarrow \mathbb{R}^n$, where $\Sigma$ is the vocabulary of values in the serialization.}

We give an example of function $\mathbf{f}$ and this example will be used in proof of proposition \ref{prop1} below. Suppose our tree has node values in set $\{w_1,w_2,\cdots,w_{M-1},w_M\}$ where $M$ is a positive integer. With pre-order serialization, the full vocabulary of for the Transformer input is $\Sigma=\{'(',')',padding\}\cup \{w_1,w_2,\cdots,w_{M-1},w_M\}$ so we have $|\Sigma|=M+3$. 

Now we define function $\mathbf{f}$ by its univariate counterpart $f:\Sigma\rightarrow\mathbb{R}$ as
\[f(x) =
\left\{
	\begin{array}{ll}
		1  & \mbox{if } x = `(` \\
		4 & \mbox{if } x = `)` \\
		9 & \mbox{if } x \in \{w_1,w_2,\cdots,w_{M-1},w_M\} \\
		0 & \mbox{if } x = \text{padding}
	\end{array}
\right.\]
For example, if the pre-order serialization of a tree is $$ ( w_1 ( w_2 w_3 w_4 ) w_5 ) $$ and $n=14$, the serialized (padded) vector is $$\mathbf{z}=['(',w_1,'(',w_2,w_3,w_4,')',w_5,')',$$ $$padding,padding,padding,padding,padding]$$ and thus the backbone vector is $$\mathbf{f}(\mathbf{z})=[1, 9, 1, 9, 9, 9, 4, 9, 4, 0, 0, 0, 0, 0].$$

Before going to the proposition, let us make some assumptions:
\begin{enumerate}
    \item Input tree is not empty.
    \item Input vector has length $n$.
    \item The serialization method of all input trees is the same.
    \item Backbone vectors cannot be learned directly in the embedding layer.
    \item At least 3 different elements are present in every backbone vector.
\end{enumerate}

The fourth assumption is reasonable since it is hard to learn a tree decoder in one linear layer. The last assumption is generally true because we expect at least one element to present padding, one to present node values, and one to give information about structures. 

We also need to clarify notations. For vector and matrix indices we use Python indexing notations. Briefly, let $\mathbf{p}$ be a vector of length $n$ and let $k$ be a positive integer, then $\mathbf{p}[-k] = \mathbf{p}[n-k]$ and $\mathbf{p}[-k:]$ selects the last $k$ entries in $\mathbf{p}$. The index starts from 0. We use bold font of a number to denote a vector with every element being that number, e.g. $\mathbf{0}=[0,0,\cdots,0]$. If there is a subscript to a bold number, it stands for the dimension of the vector. If not, the dimension should be clear by context.

We now state and prove our proposition \ref{prop1}.

\proposition{There exists a one-layer, one attention head Transformer that can find the backbone vector for a tree.}\label{prop1}

\proof{We prove this proposition by construction.  We need to show that there exists a parameter configuration that finds the backbone for the input tree. 

Let the embedding dimension be $$d=\max(n+2, [\sqrt{n\cdot (M+3)^n}]+1).$$ 
Let the embedding of any token in $\Sigma$ be a $d$-dimensional vector such that the last $n$ elements are 0. This can be easily achieved by padding any $d-n$-dimensional embedding with $n$ zeros. For example, the embedding of the first token in our input is $X_0[0,:]$ so we have $X_0[0,-n:]=\mathbf{0}$. Let the positional embedding matrix $P$ be all zeros except $P[k,-(n-k)]=1$, for $k=0, 1,\cdots,n-1$. That is, $P$ is an identity matrix concatenating to the right of a zero matrix. This way we have the first row of $X_1$ with every element equal to $X_0[0,:]$ except $X_1[0,-n]=1$. We set other trainable parameters as following:

\begin{itemize}[noitemsep,nolistsep]
\item Define $W^Q\in\mathbb{R}^{d\times d}$ be a solution for equations $(X_1 W^Q X_1^T)[0,:]/\sqrt{d}=\mathbf{y}$ for every possible $X_1$, where $\mathbf{y}$ is an n-vector such that $$\text{softmax}(\mathbf{y})=\frac{\mathbf{f}(\mathbf{z})}{\mathbf{1}_n\mathbf{f}(\mathbf{z})}, $$
where each $\mathbf{z}$ is an input vector corresponding to each $X_1$.
\item $W^K=I^{d\times d}$.
\item $W^V\in\mathbb{R}^{d\times d}$ is a matrix with all 0s except $W^V[k,k]=1$ for $k=d-n+1,\cdots,d-1$ and $W^V[k, d-n]=-1$ for $k=d-n+1,\cdots,d-1$.
\item $W_1\in\mathbb{R}^{d\times d}$ is a matrix with all 0s except 
$W_1[k,k]=1$ for $k=d-n,d-n+1,\cdots,d$. $W_2=I^{d\times d}$, $b_1[:-n]=\mathbf{0}$, $b_1[-n:]=\mathbf{9}_n$, $b_2=\mathbf{0}_d$. 
\end{itemize}

The vector $\mathbf{y}$ is well-defined because elements in  $\frac{\mathbf{f}(\mathbf{z})}{\mathbf{1}_n\mathbf{f}(\mathbf{z})}$ sum up to $1$ and all the elements are nonnegative since the input tree is not empty. 
Such $\mathbf{y}$ exists since we only need to keep the differences of elements in $\mathbf{y}$, that is, as long as $e^{y_i-y_j}=f(z_i)/f(z_j)$ holds for all $i,j\in[n]$ and $i\neq j$, this euqation holds. There are many such $y\in \mathbb{R}^n$ and we only need one of them for each $\mathbf{z}$.

Next, we show $W^Q$, as a solution to the linear system, exists. Since we only have $|\Sigma|=M+3$ different embeddings and an $n$-dimensional input, there are no more than $(M+3)^n$ possible $X_1$. Each one of them gives $n$ linear equations since $\mathbf{y}$ is an n-dimensional vector. Thus, the total number of equations is no more than $n\cdot (M+3)^n$. 
The variables in these equations are elements in $W^Q$ so there are $d^2$ of them. 
With $d\geq [\sqrt{n\cdot (M+3)^n}]+1$, these equations have a solution.

With that in mind, let us start from the beginning. In the embedding layer, we assign $X_1\in\mathbb{R}^{n\times d}$ so that $X_1[:,:-n]$ indicates the value and $X_1[:,-n:]$ indicates the position. Secondly, in self-attention layer, we know $$(\text{softmax}(\frac{QK^T}{\sqrt{d}})X_1)[0,-n:]= \frac{\mathbf{f}(\mathbf{z})}{\mathbf{1}_n\mathbf{f}(\mathbf{z})}.$$ 
After multiplying with $W^V$, we have $$(\text{softmax}(\frac{QK^T}{\sqrt{d}})V)[0,-n+1:]$$ unchanged and  $$(\text{softmax}(\frac{QK^T}{\sqrt{d}})V)[0,-n]=(\mathbf{f}(\mathbf{z})[0]-\mathbf{1}_n\mathbf{f}(\mathbf{z}))/(\mathbf{1}_n\mathbf{f}(\mathbf{z})),$$ while other elements in $$(\text{softmax}(\frac{QK^T}{\sqrt{d_k}})V)[0,:]$$ are 0. After add and norm, we get
\begin{equation*} \label{eq1}
\begin{split}
& X_2[0,-n:] \\
& = LayerNorm(X_1[0,-n:]+\frac{\mathbf{f}(\mathbf{z})}{\mathbf{1}_n\mathbf{f}(\mathbf{z})}-[1,0,\cdots,0]) \\
 & = LayerNorm([1,0,\cdots,0]+\frac{\mathbf{f}(\mathbf{z})}{\mathbf{1}_n\mathbf{f}(\mathbf{z})}-[1,0,\cdots,0]) \\
 & = LayerNorm(\frac{\mathbf{f}(\mathbf{z})}{\mathbf{1}_n\mathbf{f}(\mathbf{z})}) = LayerNorm(\mathbf{f}(\mathbf{z})).
\end{split}
\end{equation*}
Last, in feed-forward layers, consider the first row of $X_2$, i.e. $X_2[0,:]$. After multiplying with $W_1$, we have 
$$(X_2[0,:]W_1)[:-n]=\mathbf{0}, $$ $$(X_2[0,:]W_1)[-n:]=X_2[0,-n:]=LayerNorm(\mathbf{f}(\mathbf{z})).$$ 
Since $LayerNorm(\mathbf{f}(\mathbf{z}))+\mathbf{9}_n$ is always nonnegative, the ReLU function in this layer does not do anything. After add and norm, we have \begin{equation*}
\begin{split}
& X_3[0,-n:] \\
& =LayerNorm(LayerNorm(\mathbf{f}(\mathbf{z}))+LayerNorm(\mathbf{f}(\mathbf{z}))) \\
& =LayerNorm(\mathbf{f}(\mathbf{z})).
\end{split}
\end{equation*} Since we have at least 3 different elements in $\mathbf{f}(\mathbf{z})$, we can distinguish the backbone of the tree from the ratio of differences between elements. \qed
}

In fact, we can extend this proof to any other serializations of trees. There is no change in the construction and proof. In practice, if we have certain limits in tree transduction, for example, every node only cares about its parent node, then the attention matrix can be compressed by proper masking.

From the backbone, we can easily deal with some challenging tree transduction tasks. For example, if we want to find the parent of a node, we can put the parent index of every node in a vector for every backbone and the Transformer can decide which vector to take since it has the backbone of the input tree.

This proposition, on the other hand, also answers the question about how deep a tree a Transformer can handle. As long as the serialization of the tree is no longer than the input length limit $n$, our Transformer in the proof can find its structure.

\section{Expansion of Tree Backbones}
\label{sec:secondapproach}

In \autoref{sec:initialapproach}, we proved that Transformer networks can learn tree structures. All backbone vectors up to a certain depth are useful for identifying a $W^Q$. 
However, without this extra information used in $W^Q$, it is not obvious how the Transformer will learn the backbone matrix by itself. 
In this section, we want to prove that two linear layers are sufficient to learn all backbone vectors, given only a small amount of them in the first training example in the input of the first layer.

Since backbone vectors are in n-dimensional space $\mathbb{R}^n$, we first show that, surprisingly, only 2 backbones are necessary to get a basis of the whole $\mathbb{R}^n$ space using a single linear layer with ReLU activation function. This construction achieves a new basis by eliminating elements one by one.

\lemma{There exist 2 vectors in $\mathbb{R}^n$ such that one can get a basis of the whole vector space ($\mathbb{R}^n$) after going through a linear layer and ReLU function.}\label{lemma1}

\proof{Let $\mathbf{x}$ and $\mathbf{y}$ be two vectors in $n$-dimensional space, where $\mathbf{x}=(1, 1, \cdots, 1, 1)$, $\mathbf{y}=(1, 2, \cdots, n-1, n)$. For $m\in\{0,1, \cdots, n-1, n\}$, let $\mathbf{z_m}=ReLU(\mathbf{x}-\frac{1}{m+1}\mathbf{y})$. We have $z_0=\mathbf{0}$. If $m>0$, for $k=1,2,\cdots,m$, the $k$-th element of $z_m$ is
$$ \left\{
\begin{array}{cc}
\frac{m-k+1}{m+1} & \text{ for }  k=1,2,\cdots,m\\
0 & \text{ for }  k=m+1,\cdots,n 
\end{array}
\right.
$$
These $\mathbf{z}$ vectors then span the whole space.\qed}

In Lemma \ref{lemma1} we showed that with the help of the ReLU activation function we need only 2 vectors to form an entire basis for n-dimensional space.  In our proof, we explicitly provided those two vectors and proved the lemma by construction. The question that remains is if any arbitrary two backbones can serve to generate the entire space, and what are the assumptions on those backbones.  It turns out that the necessary assumptions are weaker than one might assume as we only need any two linearly independent nonzero vectors.

\theorem{In n-dimensional space, any two linearly independent nonzero vectors get a basis of the whole space($\mathbb{R}^n$) after going through two linear layers and ReLU function.}

\proof{Let us call these two vectors $v_1, v_2$. According to the previous lemma, we only need to find a $n\times n$ matrix M so that $Mv_1=\mathbf{1}$, $Mv_2=(1, 2, \cdots, n-1, n)$. In the lemma we have a second linear layer which can be thought of as left multiplying the input matrix $[\mathbf{x};\mathbf{y}]^T$, with a $n\times 2$ matrix. Here $M$ is the matrix with all the parameters needed in the first layer and with $M$ we can find the desired input for the second layer.

In order to do this, we can search for rows in $M$ one at a time. For the k-th row vector $M_k$, we need $M_kv_1=1, M_kv_2=k$, $k=1,2,\cdots,n$. Since $v_1, v_2$ are nonzero and linearly independent, such $M_k$ exists for any $k$.\qed}

Since we can achieve the same operations in Transformer networks with linear operation in linear layer and ReLU function in FFN layers, it is possible that we learn all backbone vectors in one Transformer encoder layer. As shown in the previous section, we can find tree structures by one encoder layer, so putting this together, two encoder layers are able to find tree structures efficiently. 

Now that we demonstrated the expressive power of the non-linearity, we can see that even at the start of the second layer of the Transformer network, all tree backbones that can be encoded in the $n$-dimensional vector space can be found. This implies that even with limited examples, the Transformer can learn to generalize to tree structures of an arbitrary depth! With this theorem proved, it is possible to learn parameters in $W^Q$ in \autoref{sec:initialapproach} faster with possible tree backbone vectors in the output of the encoder in this section. However, the implication that it is {\it possible} to learn does not imply that Transformers will indeed learn the tree structures if there is no explicit advantage in training. Thus, we create explicit synthetic data in order to determine whether 2-layer Transformer networks will learn tree structures.
\section{Experiments}
\label{sec:experiments}

\begin{table*}[t]
\footnotesize
\centering
\begin{tabular}{c c c c c c}
\hline
Experiment & Encoding Type & Average Corrected Acc ($\%$) & Std & Average Steps to Converge & Std\\
\hline
copy & sequential & 100 & 0 & 7440 & 219.089 \\
copy & tree & 100 & 0 & 8280 & 216.795\\
depth0 & sequential & 100 & 0 & 24150 & 946.925 \\
depth0 & tree & 100 & 0 & 11440 & 847.345\\
parent1 & sequential & 99.972 & 0.0281 & 62400 & 10032.198 \\
parent1 & tree & 100 & 0 & 50180 & 5558.507\\
parent12 & sequential & 13.161 & 1.908 & 27900 & 12330.653 \\
parent12 & tree & 14.364 & 1.937 & 55367 & 32312.149\\
parent123 & sequential & 20.409 & 15.860 & 50120 & 33506.373 \\
parent123 & tree & 14.257 & 2.451 & 50280 & 39009.383\\
sibling1l & sequential & 99.604 & 0.444 & 37100 & 1326.650 \\
sibling1l & tree & 99.122 & 1.856 & 17920 & 1042.593\\
sibling1r & sequential & 99.986 & 0.00715 & 43940 & 3126.979 \\
sibling1r & tree & 99.149 & 1.576 & 31040 & 14180.021\\
\hline
\end{tabular}
\caption{Corrected accuracy and convergence statistics for the synthetic data experiments. Each combination, of one of the seven transduction tasks (experiment) and one of two encoding types, has five runs.}
\end{table*}

In order to evaluate our theoretical results in sections 2 and 3, we ran several simulations. We designed synthetic tasks that require the standard sequential positional encoding Transformers to understand tree structures for successful completion. These simulations were designed to provide strong complementary evidence to the existing heuristic evidence showing that Transformers can learn tree structures. 

We generated a common set of trees with a maximum depth of 5 and a vocabulary size of 256 with internal nodes in the range of [0-10] and leaf nodes in the range of [0-255]. These are passed through designed deterministic context-aware tree transduction rules to generate valid output trees.  We then split this data into training and validation sets. \par

We ran each experiment five times on V100 GPU with OpenNMT toolkit \citep{klein-etal-2017-opennmt}, batch size 1024, Adam optimizer \citep{kingma2017adam}, learning rate 0.1, and sequence length 200 for 100,000 steps.

\subsection{Synthetic Data Generation}

\begin{figure}[ht]
\centering
    \includegraphics[width=0.5\textwidth]{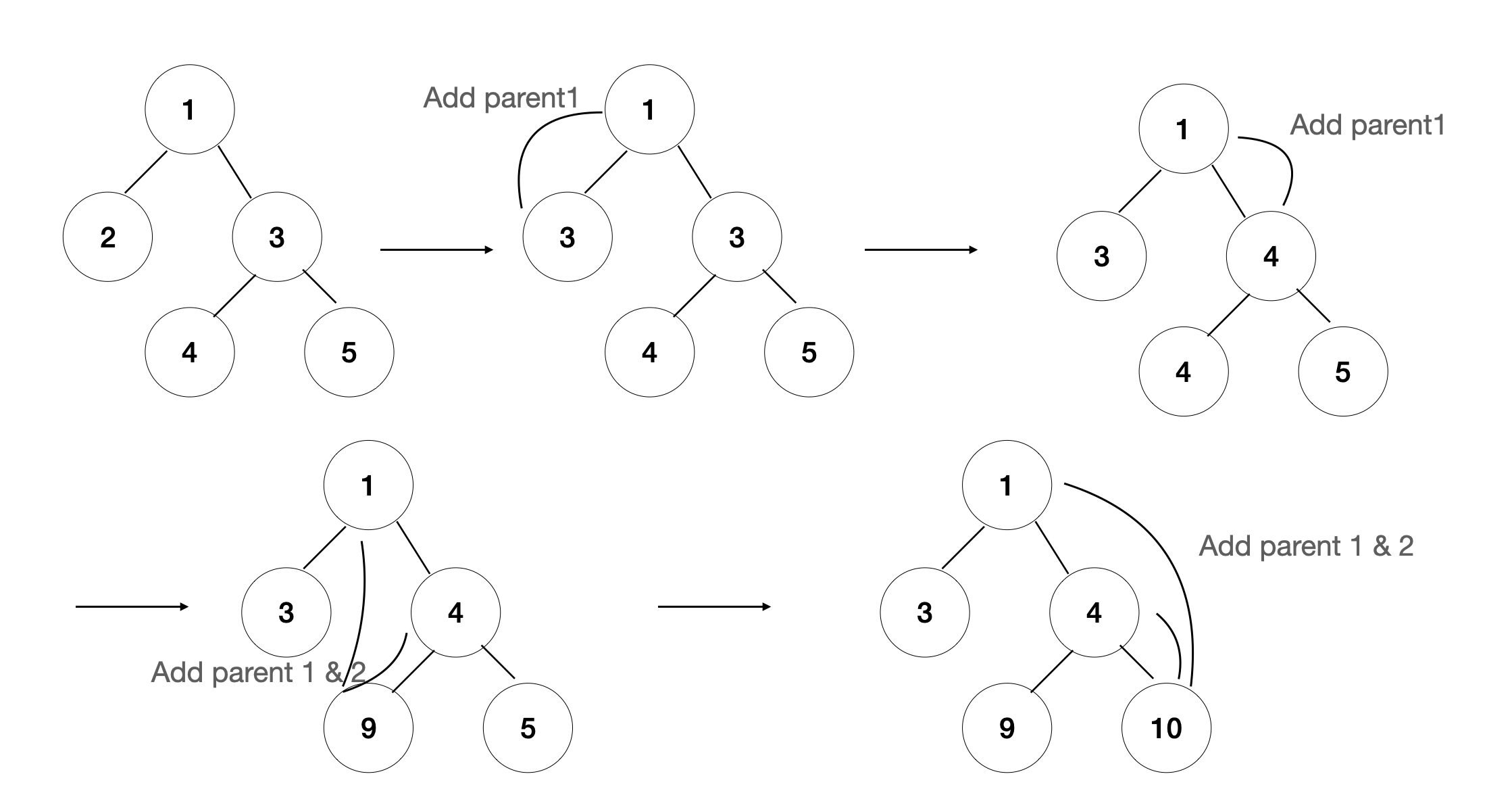}
    \caption{Step-by-step process of a 'parent12' example}
    \label{fig:exp2}
\end{figure}

We experimented with different transduction rules with varying degrees of difficulty. The rules can be derived from their set names.  For instance: 
\begin{itemize}
    \item \emph{`copy'} means the output tree is a copy to the input tree;
    \item \emph{`parent1'} means that every node value in the output tree is its input value plus its parent node's value in the input tree;
    \item \emph{`parent12'} means that nodes values in the output tree are the input value plus the value of the parent node plus the value of the grandparent node;
    \item \emph{`sibling1l'} means that every node in the output tree is the input node value plus the left child value of its sibling;
    \item \emph{`sibling1r'} means that every node in the output tree is the input node value plus the right child value of its sibling.
\end{itemize}    
The rest are similar. If some of the addends are missing (e.g. the node does not have a grandparent in case \emph{`parent12'}), we simply ignore that addend. Importantly, operations are performed in the order of the serialization.  In other words, the latter nodes may use parent node values (for instance) that have been changed from their input tree value. We illustrate a few easy examples here:
\begin{itemize}[noitemsep,nolistsep]
    \item Copy: input `( 1 2 3  )'; output `( 1 2 3  )';
    \item Parent1: input `( 1 2 3 )'; output `( 1 3 4 )';
    \item Parent12: input `( 1 2 ( 3 4 5 ) )'; output `( 1 3 ( 4 9 10 ) )'.
\end{itemize}
For clarity, we explicitly step through the 'parent12' example in \autoref{fig:exp2}.

\subsection{Positional Encodings}

As an upper bound, we include a tree positional encoding Transformer (TP-Transformer) where tree positions are explicitly encoded and thus are not learned during training. The TP-Transformer must still learn the transduction rules. With all information about tree structure given, the tasks should be easy for the Transformer, and thus this serves as a suitable baseline for isolating how well a Transformer can learn tree structure given only sequential positional encodings.

In \citet{shiv2019novel}, the authors created a clever indexing method that incorporates tree-structured positional encodings which maintain linearity and can be applied to the Transformer.
Here we adapted the idea to emphasize the importance of tree position by repeating tree-structured positional encodings many times so as to cover as many dimensions as possible.
For example, if the encoding in \citet{shiv2019novel} is $\mathbf{p}=(p_1, p_2, \cdots, p_k, 0, \cdots, 0)$, our encoding is $\mathbf{p}=(p_1\mathbf{1}_n, p_2\mathbf{1}_n, \cdots, p_k\mathbf{1}_n, 0, \cdots, 0)$ where $n\in\mathbb{Z}^+$ is chosen so that $n*k$ is as large as possible without exceeding the dimension of $\mathbf{p}$.

Now we explain how the positional encodings $\mathbf{p}$ are generated. Suppose $n$ is the degree of the tree and $k$ is the max tree depth, then $\mathbf{p}$ has dimension $n\cdot k$. Root position is encoded as $\mathbf{0}$ while others are encoded based on its path from the root. If the path is $<b_1,\cdots,b_L>$, where $b_i$ is the step choice, left or right, at $i$th layer and $L$ is the layer of this node, then for this node $x$, the positional encoding is $x=D_{b_L}D_{b_{L-1}}\cdots D_{b_1}\mathbf{0}$, where $D_i$ is an operation defined as $D_i x=e_i^n;x[:-n]$. Here $e_i^n$ is a one-hot $n$-vector with hot bit $i$ and $;$ means concatenation.

In the TP-Transformer, we make the original input tree full by adding nominal nodes with value $-1$ and serialize it by taking all the node values into a sequence, top to bottom and left to right. For example tree $(1 2 (3 4 5))$ becomes $[1, 2, 3, -1, -1, 4, 5]$ as an input to the TP-Transformer. This formulation avoids encoding tree structures in the embedding layer but still finds tree positional encodings easily.

\subsection{Evaluation and Discussion}

\begin{figure}[ht]
\centering
    \includegraphics[width=0.5\textwidth]{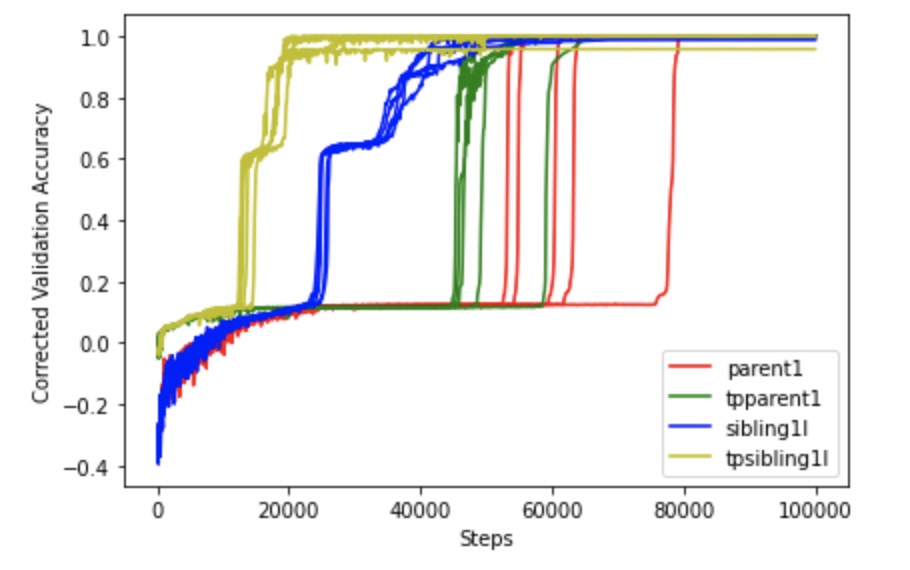}
    \caption{Performance of two types of positional encodings. Transformers with tree positional encodings converge faster than ones with sequential positional encoding.}
    \label{fig:exp1}
\end{figure}

\begin{figure}[ht]
\centering
    \includegraphics[width=0.5\textwidth]{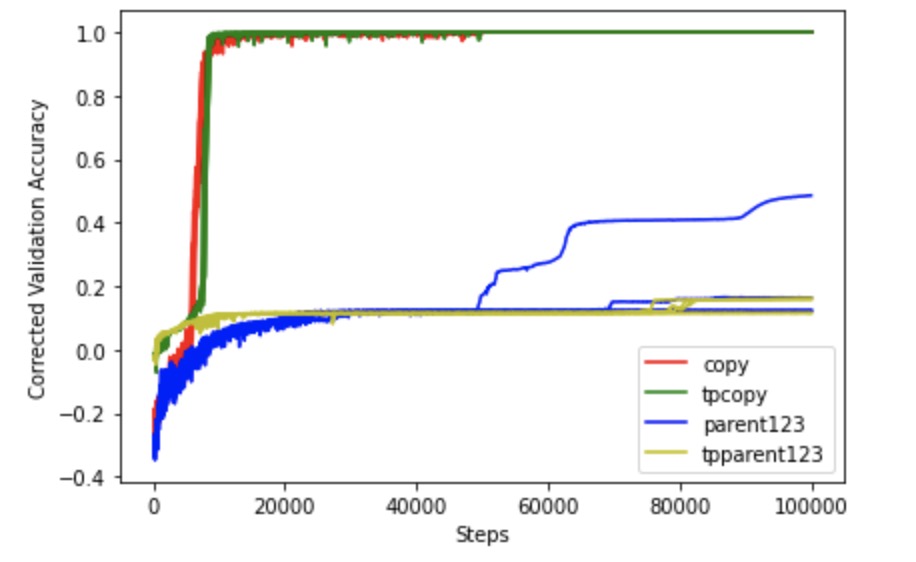}
    \caption{Performance of two types of positional encodings in extreme tasks.}
    \label{fig:exp3}
\end{figure}

We ran each experiment five times for each transduction task, considering both using sequential positional encodings and using tree positional encodings. The training objective is cross-entropy loss.

A simple approach to evaluate the two methods is to calculate the accuracy based on the number of correct tokens in the output sequence. However, Transformers always correctly identify non-value elements, but those are less important for our purposes. Thus, in order to better compare two encoding methods, we use a corrected accuracy. The corrected accuracy involves calculating the ratio of non-value elements in each task and assuming Transformers get these correct. For example, in '\emph{parent1}' task $51.3\%$ of the tokens in the input are non-value, so we have 
\[
corrected\ accuracy=\frac{original\ accuracy-48.7}{51.3}.
\]
Corrected accuracy can be negative at first, but it is a fair criterion to compare the final performances of two encoding methods.

We summarize the results in Table 1. Here, we use the corrected validation accuracy and we define converging as attaining $99\%$ of corrected final accuracy. Generally speaking, Transformer networks converge quickly on low complexity transduction rules like '\emph{copy}', converges slower but still solve medium complexity problems like '\emph{sibling1l}', but never learns hard problems like '\emph{parent12}'.

In most tasks, experiments with tree positional encodings converge faster than ones with sequential positional encodings. Some of the tasks, e.g. parent12 and parent123, are challenging since they require information from more nodes and those nodes are further away in the tree as shown in \autoref{fig:exp2}. In the harder experiments where corrected final accuracy is substantially less than $100\%$, two encodings have similar performance. Importantly, despite the fact that TP-Transformers converge in fewer steps, the actual runtime is longer due to longer inputs to TP-Transformers.  For instance, a parent1 (a sequential Transformer on the parent1 task) run takes 28761 seconds while a tpparent1 (a TP-Transformer on the parent1 task) takes 50181 seconds.

We show the training processes of task \emph{parent1} and \emph{sibling1l} in \autoref{fig:exp1}, where the lines are different runs of the same task. In the legends, 'tp' means tree positional encodings while no 'tp' means sequential encodings. \autoref{fig:exp1} shows that in both tasks tree positional encodings have faster convergence and smaller variance. 

In \autoref{fig:exp3} we show performance on the extreme ends of task difficulty: the easiest task \emph{copy} and hardest task \emph{parent123}. The purpose of the \emph{copy} task is to compare performance when tree information is not necessary.  The simple strategy works better in this task, as expected. The \emph{parent123} task compares performances in a task in which both methods learn little about the relationship, even if tree structure information is given directly. Notice that there is one run where '\emph{parent123}' has higher corrected accuracy. Thus, Transformers are capable of learning those hard relationships.  We believe that results may improve and become more stable with a deeper network or more steps. Because we are only comparing two positional encoding approaches here, we do not investigate this further.

These experiments show that Transformers with either form of positional encodings can learn tree transduction tasks.  As expected, explicit tree encodings converge more quickly and to higher accuracy.  In experiments where accuracy is much lower than $100\%$, the results of both encoding methods are close to random guessing. This indicates that standard sequential positional encoding Transformers can learn tree structure relationships, but extra information may be necessary to improve performance in complicated relationships. These results agree with our theoretical conclusions in sections 2 and 3.

\section{Related Work}
\label{sec:relatedwork}

Prior empirical work shows that Transformer networks and other neural networks learn tree structures during training. Several works examine if and how tree structures are captured by neural networks, e.g., \citet{hewitt2019structural} find that syntax trees  are embedded in linear transformations of hidden state layers of both ELMo and BERT. \citet{jawahar2019does} discover that BERT representations capture tree-like structure in linguistic information. \citet{lin2019open} conclude that BERT does model linguistic hierarchical structure, but is less sensitive to reflexive anaphora. \citet{marevcek2018extracting} try to find tree structure from attention information in Transformers.
Another research avenue is to explicitly integrate tree structures into self-attention sections in the Transformer~\citep{Wang2019TreeTransformer,omote-etal-2019-dependency,duan2019syntax,wang2019selfattention}. Similarly, \citep{harer2019tree} added Tree Convolution Block to promote the performance of the Transformer on tree tasks. All of these works empirically examine either the discovery of the usefulness of encoding tree structures in Transformer networks; however, none attempt to understand theoretically why these structures are encoded.

One of the reasons why people are interested in learning tree structures is 
that tree representations of natural languages are common in the linguistics community~\citep{santorini2007syntax}. Dependency and constituency trees allow for better treatment of both morphologically rich languages and ones with free word order \citep{berg2011structure}. The locality of semantic information in tree models is captured at a higher resolution \citep{Wang2019TreeTransformer}. In short, tree representations can improve the performance in general natural language processing (NLP) tasks such as machine translation and text simplification \citep{shu2019generating, paetzold2013text}. Because of the benefits from leveraging tree representations, large subfields of NLP focus on tree representation methods, like dependency parsing \citep{Zhang2015-bg} and syntax-based (tree-to-tree) machine translation models \citep{cowan2008tree,razmara2011application}. 

On the other hand, researchers have examined the computational power of neural nets. Many neural network structures have been proved to have Turing completeness, as \citet{siegelmann1995computational} proved Turing completeness of recurrent neural networks, even with bounded neurons and weights, and recent work, \citet{perez2021attention}, showed Turing completeness of the Transformer with one encoder layer and three decoder layers. With limited precision, in reality, neural networks have also been an approximation of important functions. \citet{yun2019transformers} proved that Transformer models with positional encodings can universally approximate any continuous sequence-to-sequence functions on a compact domain. 
In \citet{zaheer2020big}, through a sparse attention mechanism, a universal approximation of sequence functions has been proposed with linear dependency on sequence length while keeping Turing completeness. All of them showed, in different ways, the computational power of the Transformer in general.

To further the previous work, we analyze the theoretical limitations of the representation of trees by Transformers.
Previous work proved the importance of some kind of grounding in learning, e.g., \citet{merrill2021provable} found ungrounded language models are limited in ability to understand in the way that models have a problem understanding non-transparent patterns in language. As backbones represent tree structures, we use a Transformer to find tree backbones with one single encoder layer. Some form of grounding is essential for models to learn. Although tree backbones can be found in the Transformer based on its Turing completeness in \citet{perez2021attention}, here we only need two encoders, rather than one encoder and three decoders in the general case, and we do not assume hard-attention. 
Thus, our work concludes that a simple 2-layer one attention head Transformer can learn tree structures.


\section{Conclusion}


In this paper, we discussed the theoretical possibility of representing tree structure in the Transformer. In proposition 1, we proved we can find the backbone vector of any input tree in a one-layer Transformer. In theorem 1, we learned that 2 linearly independent vectors and two linear layers with ReLU activation are sufficient to learn all tree backbones. Combining these, we proved that we can always understand a tree with a 2-layer Transformer. 

Synthetic data experiments showed empirically that Transformers do learn tree structures while in most tasks, explicit tree information decreases the training steps necessary. We expect that further (computationally intensive) experiments can lead to a clearer asymptotic boundary of how much tree information can benefit the training process. 

Therefore, we now have answers to questions raised in the introduction. Transformers can capture tree structures. Even a 1-layer 1 attention head Transformer can deal with an input tree whose serialization is no longer than its input length limit. However, in \autoref{sec:secondapproach}, we find that it is necessary to have a 2-layer 1 attention head Transformer to learn the backbones of any possible serialization. Finally, we can make the learning process more efficient by modifying the Transformer network to explicitly encode tree structure information.

In summary, our theoretical findings support the empirical results that many NLP tree transduction problems can be solved using Transformer networks. Our empirical results show that Transformers are able to learn tree structures.

\bibliographystyle{aaai22}
\bibliography{t2tl_ref}

\end{document}